\documentclass[10pt,twocolumn,letterpaper]{article}

\usepackage{iccv}
\usepackage{times}
\usepackage{epsfig}
\usepackage{graphicx}
\usepackage{amsmath}
\usepackage{amssymb}
\usepackage[linesnumbered,ruled,vlined]{algorithm2e}
\usepackage{booktabs}       

\SetKwInput{KwInput}{Input}                
\SetKwInput{KwOutput}{Output}              
\SetKwInput{KwInit}{Initialization}         

\usepackage[breaklinks=true,bookmarks=false]{hyperref}

\iccvfinalcopy 

\def\iccvPaperID{****} 

\ificcvfinal\fi

\begin{document}

\title{Learning to Detect and Retrieve Objects from Unlabeled Videos}

\author{%
  Elad Amrani$^{1,2}$, Rami Ben-Ari$^{1}$, Tal Hakim$^{1}$, Alex Bronstein$^{3}$ \\
  $^{1}$IBM Research AI \\
  $^{2}$Department of Electrical Engineering, Technion, Haifa, Israel\\
  $^{3}$Department of Computer Science, Technion, Haifa, Israel\\
  \tt\small {elad.amrani@ibm.com},
  \tt\small {ramib@il.ibm.com},
  \tt\small {thakim@il.ibm.com},
  \tt\small {bron@cs.technion.ac.il}
}

\maketitle
\ificcvfinal\thispagestyle{empty}\fi

\begin{abstract}
Learning an object detector or retrieval requires a large data set with manual annotations. Such data sets are expensive and time consuming to create and therefore difficult to obtain on a large scale. In this work, we propose to exploit the natural correlation in narrations and the visual presence of objects in video, to learn an object detector and retrieval without any manual labeling involved. We pose the problem as weakly supervised learning with noisy labels, and propose a novel object detection paradigm under these constraints. We handle the background rejection by using contrastive samples and confront the high level of label noise with a new clustering score. Our evaluation is based on a set of $11$ manually annotated objects in over 5000 frames. We show comparison to a weakly-supervised approach as baseline and provide a strongly labeled upper bound.
\end{abstract}

\begin{figure*}
\begin{center}
\includegraphics[width=\textwidth,keepaspectratio]{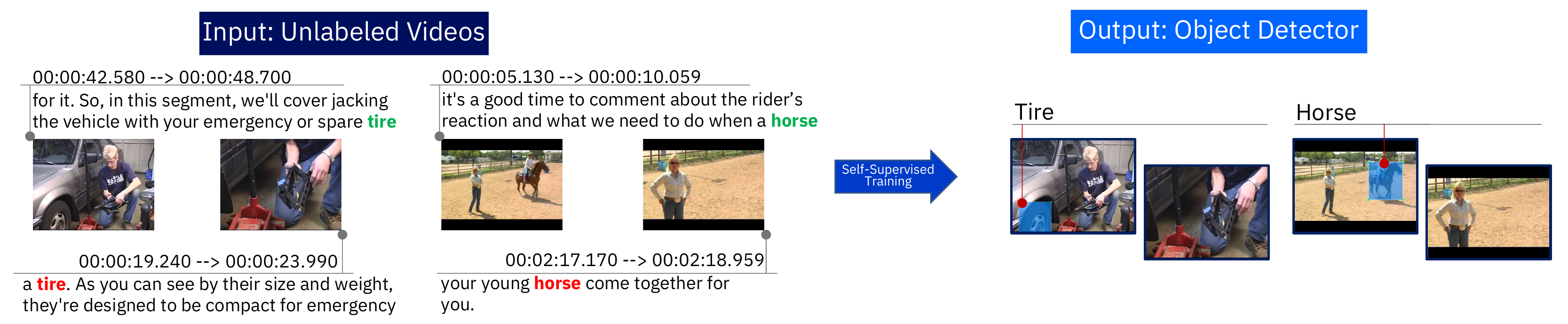}
\end{center}
   \caption{\textbf{Self-Supervised Object Detection.} We use unlabeled videos as input and train an object detector without any manual annotation. Self-supervision is based on the timed transcript selected automatically, using closed captions (or speech-to-text). For a given object name, we extract a single key frame at the instant the object was mentioned. Yet, only in part of the frames the object is visible (colored as {\color{green}green}). Often the object is mentioned but out of the frame or missing (colored as {\color{red}red}). The object can be detected by searching common themes relating the selected frames and discriminating them from frames in irrelevant videos. Zoom in for better visibility.}
\label{fig:description}
\end{figure*}

\section{Introduction}
While humans seem to learn from minimal supervision, existing machine learning techniques often require a tremendous amount of labeled data. One possibility allowing to break free from this limitation is the self-supervised learning paradigm allowing to harness huge unlabeled data sets. Existing use cases suggest reinforcement learning \cite{Reinforcement_Data_Efficient_NIPS2018}, solving a Jigsaw puzzle or image colorization \cite{GoyalarXiv2019,ZhangECCV2016}. Yet these methods require a reward function, or introduce very specific tasks. In this paper we address a different problem of Self-Supervised Object Detection and Retrieval (SSODR) from unlabeled videos. Large-scale video data sets such as YouTube-8M \cite{youtube8m} and How2 \cite{how2DataSet} can be leveraged for this purpose. Fig. \ref{fig:description} describes the targeted problem in this paper.

Given unlabeled videos, the audio channel can be used as a "free" source of weak labels. For instance, by seeing and hearing many frames where the word "guitar" is mentioned, it should be possible to detect the guitar due to its shared characteristics over the frames. Yet, self-supervised learning performed in the wild is hard, as the audio and the visual contents may often appear completely unrelated.
We therefore pose the problem as weakly labeled task with noisy labels. First, we map the audio track to text using automatic transcription from a speech to text model. We then extract candidate frames from our video data set, in this case instructional videos, where the transcript contains the object name. Since the object often will not appear in all the selected frames, this set of frames introduces a weakly and noisy labeled set. Creating a contrastive negative set, that lacks the object of interest, allows discriminating between the object and background, essentially solving the object detection problem. The task is similar to weakly supervised object detection (WSOD), yet is distinct from it in the high level of label noise. The contributions in this paper are three folds: (1) We introduce a methodology for detection and retrieval of objects, learned from continuous videos without any manual labeling. (2) We pose the problem as a weakly supervised object detection with noisy labels and suggest a novel scheme to address this problem. (3) Our scheme incorporates a novel cluster scoring that distills the detected regions separating them from the surrounding clutter.

\section{Related Work}
Synchronization between the visual and audio or text as a source for self-supervised learning has been studied before \cite{HarwathNIPS2016,Suris2018}. In \cite{HarwathNIPS2016}, the authors suggest a method to learn the correspondence between audio captions and images, for the task of image retrieval. Another line of studies target Audio-Visual alignment in unlabeled videos  \cite{LookListenAndLearn2017,Gao2018,Kobar2018,CholSoung2016,Sun2016,Yu2013,Zhao2018} capitalizing on the natural synchronization of the visual and audio modalities to learn models that discover audio-visual correspondence. In \cite{LookListenAndLearn2017} the authors also suggest an object localization, yet by activation heat maps (in low resolution) and not as detection. Grounded language learning from video has been studied in \cite{CholSoung2016,Sun2016,Yu2013} in a supervised way, by learning joint representations of sentences and visual cues from videos.
In this paper we address a different problem of {\it self-supervised object detection and retrieval} from unlabeled and unconstrained videos. The closest work to our study is \cite{HarwathACL2017} that attempts to map word-like acoustic units in the continuous speech to relevant regions in the image. However this method is not self-supervised as the captions are generated manually.
\section{Method}
\label{method}
\begin{figure*}
\begin{center}
\includegraphics[width=\linewidth,keepaspectratio]{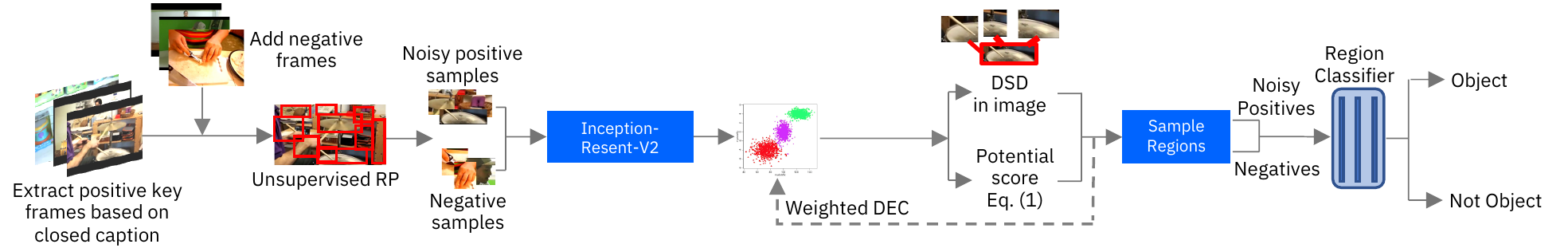}
\end{center}
  \caption{The proposed self-supervised object detection training scheme. We collect our "Positive" set according to Fig. \ref{fig:description}. "Negative" set of frames from disparate videos are added. Region proposals are generated using Selective Search and embedded into a feature space. We then cluster regions with Weighted Deep Embedded Clustering, using a potential score that allows distilling the clusters with the object of interest. A Dense Subgraph Discovery (DSD) model allows capturing high quality regions and generate our hard negative samples. Fed by the distilled regions, we train a binary classifier that separates object and non-object regions, resulting in a self-supervised object detector.}
\label{fig:ArchFigure}
\end{figure*}

The proposed self-supervised scheme is described in Fig. \ref{fig:ArchFigure}. The building blocks of our scheme are as following.

\textbf{Extraction of key frames.}
For a given object, we extract a single frame from the temporal segment containing the object's name in the transcription. Since object instances are likely to appear in these frames, we dub these frames as our ({\bf noisy}) {\it positive} set, labeled as $Y_l=1$. We introduce a balanced contrastive data set, $Y_l=0$, as our {\it negative} set, containing frames randomly selected from disparate videos, where the object was not mentioned. These frames will rarely contain the object of interest, but should include associated elements present in the surroundings of the desired object instances, in the positive set. Now we extract $N$ region proposals from the selected frames, using an unsupervised method such as Selective Search \cite{ss_ijcv}. Regions are labeled as positive / negative according to the corresponding frame label, $y_{li}=Y_l$. Using a pre-trained Inception-ResNet-v2 CNN \cite{inceptionResnetv2}, we map each region to a feature space, represented by $z_{li}$.

\textbf{Clustering and Scoring.} The purpose of our learning approach is to find a common theme across positive regions that is less likely to exist in negative counterparts. To this end, we start with clustering the regions in the embedded space. 
As region labels are not provided, we search for \textbf{clusters}, satisfying the following three conditions: (1) High purity, defined as the percentage of positive samples in the cluster (note that not all positive samples are instances of the object); (2) Low cluster variance, for tendency to include a single object type; and (3) High video variety: prioritizing clusters that include regions from a higher variety of videos, since we expect the object to have common characteristics among various videos. This property also copes with the high temporal correlation in a single video, that might create dense clusters of regions from the a long video. We formalize these constraints using the following softmax function $S_k$, to which we refer as the {\it potential score}. This score reflects the likelihood of cluster $k$ to contain the object of interest: 
\begin{equation}
    \label{score_func}
    S_k = \sigma\bigg(\tau \frac{P_k^2\cdot \log{U_k}}{V_k}\bigg)\;\;\;\;\;\;\;k\in\{0..K-1\}.
\end{equation}
Here, $\sigma(\cdot)$ is the softmax function, $K$ denotes the total number of clusters, $\tau \in \mathbb{R}$ (hyper-parameter), $P_k$ is the positive-ratio, $V_k$ is the cluster distance variance, and $U_k$ denotes the number of unique videos incorporated in the cluster. All parameters are normalized to unit sum.

\textbf{Clustering.} Following feature extraction of region proposals, we cluster the proposals using a variation of deep embedded clustering (DEC) \cite{unsupervised_dec} that uses the {\it weighted} student's t-distribution as a similarity measure:
\begin{equation}
    q_{i,j} = \frac{(1+||z_i - \mu_j||^2)^{-1} \cdot w(i, j)}{\sum_{j'}{(1+||z_i - \mu_{j'}||^2)^{-1}}  \cdot w(i, j')}
    \label{eq:qij}
\end{equation}
Here, indices $i$ and $j$ are associated with the sample and cluster respectively, $z_i$ corresponds to the region embedding and $\mu_j$ is the cluster centroid (for more details on DEC please refer to \cite{unsupervised_dec}). We drop the frame index $l$ for simplicity. The newly added weights $w(i,j)$ are set according to the region label $y_i \in \{0,1\}$:
\begin{equation}
    w(i, j) = 
    \begin{cases}
    0.5 ,& \text{if } y_i = 0 \\
    1,              & \text{otherwise}
    \end{cases}
\end{equation}
increasing the loss in clustering of positive samples. Clusters are refined with new weights every $I$ epoches. Cluster centroids are optimized, while embeddings remain fixed. 
\textbf{Region Selection.}
A frequent shortcoming of weakly supervised object detection (WSOD) approaches is in their inability to distinguish between candidates with high and low object overlap. However, for training a high-performance object detector, region proposals having high overlap with the object are required. To this end, we use the Dense Subgraph Discovery (DSD) approach, previously used for WSOD \cite{DSD_JieCVPR2017}. High overlap regions are correlated with the most connected nodes in the graph (edges are connected between overlapping regions).
Unlike \cite{DSD_JieCVPR2017}, we further make use of the remaining regions as "hard negative" examples.

\textbf{Sampling and training.}
Each cluster is assigned a potential score as defined in \eqref{score_func}. We extract positive samples (regions) from high potential scores and train our detector as a binary classifier, distinguishing between these regions that are likely to contain the object of interest, and background (that may include also other objects). In order to make the training, noise robust, we conduct the following sampling regime: Positive regions that satisfy DSD criteria are sampled according to the cluster potential score distribution while negatives are sampled uniformly from the negative frames and are combined with the rejected regions from DSD (used as hard negatives). With the improvement of cluster score, the noise level in the training is reduced, purifying the region classifier (detector) input toward a nearly "noise-free" labels. Our detector is a multilayer perceptron with three fully connected layers and cross-entropy loss. In every training cycle, we initialize the training with weights from previous iteration (see Fig. \ref{fig:ArchFigure}).
\section{Experiments}
\label{section:experiments}
\textbf{Data set.} Our evaluation is based on the How2 data set \cite{how2DataSet} that includes 300 hours of instructional videos with synchronized transcription in English. By processing the caption text of the entire data set, we extract all object nouns and choose 11 top references, avoiding objects with dual verb and noun meanings as well as objects with extreme appearance variability. However, we still include in our corpus challenging nouns such as "bike" that corresponds to both bicycle \& motorbike or "gun" that refer to pistol, rifle \& glue gun or even "cup" that corresponds to glass, porcelain, paper and measuring cup.
This results in a total of 5,120 frames, with an average of 465 frames per-object. Our transcript-based frame selection introduces $52\%$ noisily labeled frames on average (i.e., only 48\% of frames are correctly labeled and contain the object of interest), presenting 2,457 frames with the true weak labels. The statistics behind our data set are shown in Table \ref{table:DataCharacteristics}. For validation of our self-supervised scheme, we manually annotated the objects in the selected frames and performed detection, separately for each class.
\begin{table*}[t]
  \small
  \centering
  \setlength\tabcolsep{5pt}
  \begin{tabular}{lcccccccccccc}
    \toprule
    \cmidrule(r){1-2}
                    & Bike   & Cup   & Dog    & Drum   & Guitar & Gun    & Horse  & Pan  & Plate  & Scissors  & Tire  & Total/Average \\
    \midrule
    No. of frames  & 419 & 583 & 1243 & 457 & 442 & 182 & 597 & 351 & 341 & 200 & 305  & 5120\\
    Obj. instances      & 715   & 628   & 551    & 312   & 564 & 111    & 327  & 154  & 147  & 92  & 526 & 4127 \\
    Noise level \%    & 28.4  & 54.4   & 59.5  & 53.4  & 35.1  & 61.5  & 47.1  &  68.4  & 67.7  & 58.5  &  22.6 & 51.5\\
    SS recall \% &  85 & 85 & 97 & 98 & 94 & 95 & 99 & 94 & 97 & 83 & 83 & 91.8\\
    \bottomrule
  \end{tabular}
  \caption{Data set: Statistics of the selected frames. \emph{Noise level} refers to the percentage of selected frames without the object present and \emph{SS recall} to selective search recall \cite{ss_ijcv} at $IOU \geq 0.5$. }
  \label{table:DataCharacteristics}
\end{table*}

\begin{table*}[th]
  \small
  \centering
  \setlength\tabcolsep{5pt}
  \begin{tabular}{llccccccccccccccc}
    \toprule
    \cmidrule(r){1-2}
            IoU    & Method    & Bike   & Cup   & Dog    & Drum   & Guitar & Gun    & Horse  & Pan  & Plate  & Scissors  & Tire & mAP  \\
    \midrule
        &FS   & 42.7   & 41.4   & 60.0    & 51.8   & 55.4 & 49.0    & 73.2  & 43.8  & 38.7  & 32.0  & 48.6 & 48.8  \\
    0.5 &WS  & F   & 2.6   & 3.2    & F   & 6.6 & 0.2    & 11.3  & 10.9  & {\bf 13.7}  & 0.6  & 11.1 & 6.7*  \\
        &Ours  & {\bf 20.3}   & {\bf 4.0}   & {\bf 12.7}    & {\bf 28.1}   & {\bf 13.0} & {\bf 9.1}    & {\bf 25.2}  & {\bf 11.6}  & 6.3  & {\bf 11.3}  & {\bf 16.8} & {\bf 14.4}  \\
    \midrule
         &FS   & 54.4   & 48.0   & 67.1    & 58.9   & 65.7 & 60.1    & 77.8  & 50.8  & 43.5  & 38.2  & 56.9 & 56.5  \\
    0.3  &WS  & F   & 4.9   & 7.3    & F   & 21.4 & 4.1  & 35.3  & 18.1  & {\bf 20.8}  & 1.2  & 16.2 & 14.4*  \\
         &Ours  & {\bf 28.0}   & {\bf 5.9}   & {\bf 21.7}  & {\bf 35.7}   & {\bf 28.3}  & {\bf 33.0}    & {\bf 40.8}  & {\bf 27.3}  & 7.1  & {\bf 13.0}  & {\bf 21.4} & {\bf 23.9}  \\
    
    \bottomrule
  \end{tabular}
  \caption{Evaluation on \textbf{test set: Average over 3 folds} - AP \%. FS: Fully Supervised, presenting an upper bound. WS: Weakly Supervised. F: Failed to converge, *: Without the failed objects. Best results comparing to WS are in bold.}
  \label{table:Results3FoldsAvg}
\end{table*}

\textbf{Baseline and upper bound.} As our baseline, we opted for the Proposal Cluster Learning (PCL) WSOD method \cite{PengPCL_TPAMI2018} currently achieving the highest mAP of 19.6\% for WSOD on ImageNet-Det. We trained PCL, in the same manner as our SSODR, with the same input, i.e., noisy positive samples, and the negative class labeled as background.
As an upper bound reference, we report the performance of the fully-supervised version of our method, where the detector is trained with ground truth labels. These comparisons also emphasized the challenge of learning object detection from unconstrained videos, manifesting motion blur, extreme views (far-field and close-ups), and large occlusions. 

\textbf{Evaluation.} We randomly split our data into 80\%-20\% train-test sets containing mutually exclusive frames, and evaluate the performance on $3$ randomized folds. Since our task is self-supervised, we allowed frames from the same video to participate in both train and test sets.
For quantitative evaluation, we manually annotated the bounding boxes of $11$ object categories. Note that annotations were used only for the testing and were not available at training, to keep the method annotation-free. As the evaluation criterion, we use the standard detection mean average precision (mAP) with different IoU thresholds. Since the positive set is noisy (see Table \ref{table:DataCharacteristics}), the evaluation was also applied to frames without the objects as well.

\textbf{Detection results.}
Quantitative results are summarized in Table \ref{table:Results3FoldsAvg}. Our methods attains an overall mAP of 14.4\% and 23.9\% for IoU over 0.5 and 0.3 respectively. 
Applying PCL yields inferior mAP on all the objects except "Plate", due to the lack of robustness of PCL to label noise. In fact, we faced convergence failure in PCL due to the noisy labels, for Bike and Drum categories. 
Overall, for the 9 successfully learned objects, we obtained a mAP of 6.7\% vs. 12.2\% @IoU=0.5, and 14.4\% vs. 22.1\% @IoU=0.3, both favorably for our approach.
As the upper bound, we present the results from our trained detector (see Fig. \ref{fig:ArchFigure}) when fed with true regions in SS proposals. Although SSODR is still far from this strongly labeled scenario, we believe that our results and labeled data can motivate others to tackle the challenging task of SSODR.

\begin{figure}[ht]
    \centering
    \includegraphics[width=\linewidth,keepaspectratio]{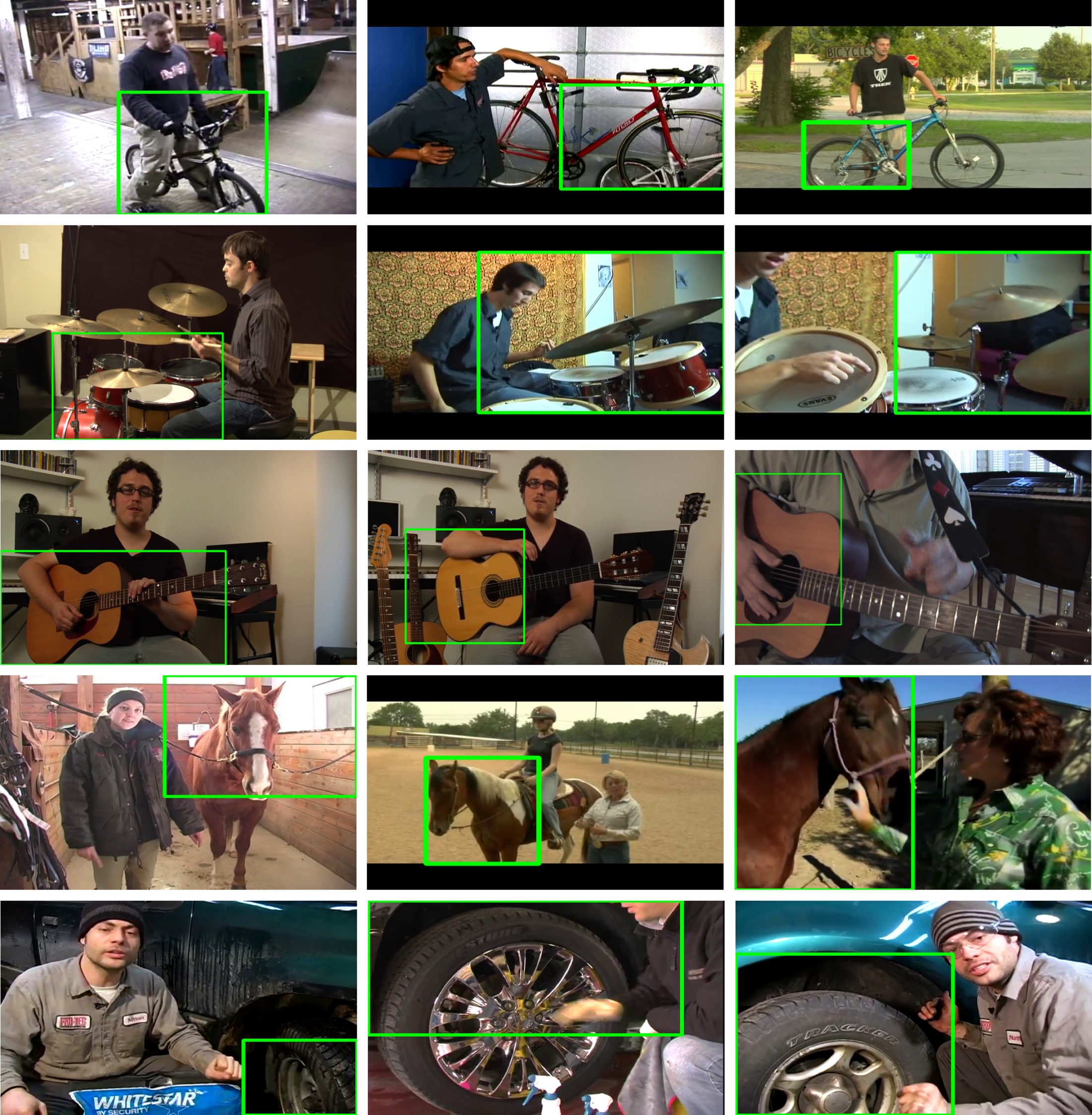}
    \caption{\textbf{Top 3 retrieval results per object}. Top to bottom: Bike, Drum, Guitar, Horse and Tire. We retrieve 3 samples from different videos that are closest to the centroid of the top scoring cluster.}
    \label{fig:retrieval}
\end{figure}

\textbf{Retrieval results.} To retrieve the top $n$ samples, given an object label we first choose the cluster with highest potential score. Filtering out regions using DSD, we extract the $n$ closest samples to the centroid. Qualitative retrieval results for 5 objects are shown in Fig. \ref{fig:retrieval}. 

\textbf{Implementation details.}
For clustering, we use $K=50$, and $\tau = 50$ in \eqref{score_func}. In our region classifier we use 3 FC layers (1024,1024,2) with a ReLU activation in layers 1-2 and a softmax activation for the output layer. Dropout is used for the two hidden layers with probability of 0.8. The classifier is trained with the cross-entropy loss function. We use ADAM for optimization with a learning rate of $10^{-4}$. The learning rate is decreased by a factor of 0.6 every 6 epochs. We train our model for 35 epochs for all objects. 
\section{Summary}
We have presented a model for the challenging task of self-supervised learning of object detection and retrieval from unlabeled videos. We pose the problem as weakly and noisily-labeled supervised learning. Our object detection is based on a model that captures regions with a common theme across selected frames. Frame selection is made according to the video transcription. This new region-level approach shows promising results in detection of objects despite their high appearance variability and multiple sub-classes arising from the language ambiguities.  Our method handles label noise in a weak setting, and is capable of learning to detect and retrieve objects without any human labeling.
{\small
\bibliographystyle{ieee}
\bibliography{egbib}

\begin{thebibliography}{10}\itemsep=-1pt

\bibitem{youtube8m}
S.~Abu-El-Haija, N.~Kothari, J.~Lee, A.~P. Natsev, G.~Toderici, B.~Varadarajan,
  and S.~Vijayanarasimhan.
\newblock Youtube-8m: A large-scale video classification benchmark.
\newblock In {\em arXiv:1609.08675}, 2016.

\bibitem{LookListenAndLearn2017}
R.~Arandjelovic and A.~Zisserman.
\newblock Look, listen and learn.
\newblock In {\em ICCV}, 2017.

\bibitem{Gao2018}
R.~Gao, R.~Feris, and K.~Grauman.
\newblock Learning to separate object sounds by watching unlabeled video.
\newblock In {\em CVPR}, 2018.

\bibitem{GoyalarXiv2019}
P.~Goyal, D.~Mahajan, A.~Gupta, and I.~Misra.
\newblock Scaling and benchmarking self-supervised visual representation
  learning.
\newblock In {\em arXiv:1905.01235}, 2019.

\bibitem{HarwathACL2017}
D.~Harwath and J.~Glass.
\newblock Learning word-like units from joint audio-visual analysis.
\newblock In {\em ACL}, 2017.

\bibitem{HarwathNIPS2016}
D.~Harwath, A.~Torralba, and J.~R. Glass.
\newblock Unsupervised learning of spoken language with visual context.
\newblock In {\em NIPS}, 2016.

\bibitem{DSD_JieCVPR2017}
Z.~Jie, Y.~Wei, X.~Jin, J.~Feng, and W.~Liu.
\newblock Deep self-taught learning for weakly supervised object localization.
\newblock In {\em CVPR}, 2017.

\bibitem{Kobar2018}
B.~Korbar, D.~Tran, and L.~Torresani.
\newblock Cooperative learning of audio and video models from self-supervised
  synchronization.
\newblock In {\em NeurIPS}, 2018.

\bibitem{Reinforcement_Data_Efficient_NIPS2018}
O.~Nachum, S.~Gu, H.~Lee, and S.~Levine.
\newblock Data-efficient hierarchical reinforcement learning.
\newblock In {\em NeurIPS}, 2018.

\bibitem{how2DataSet}
R.~Sanabria, O.~Caglayan, S.~Palaskar, D.~Elliott, L.~Barrault, L.~Specia, and
  F.~Metze.
\newblock {How2:} a large-scale dataset for multimodal language understanding.
\newblock In {\em Proceedings of the Workshop on Visually Grounded Interaction
  and Language (ViGIL)}. NeurIPS, 2018.

\bibitem{CholSoung2016}
Y.~C. Song, I.~Naim, A.~Al~Mamun, K.~Kulkarni, P.~Singla, J.~Luo, D.~Gildea,
  and H.~Kautz.
\newblock Unsupervised alignment of actions in video with text descriptions.
\newblock In {\em Proceedings of the Twenty-Fifth International Joint
  Conference on Artificial Intelligence}, IJCAI'16, pages 2025--2031. AAAI
  Press, 2016.

\bibitem{Sun2016}
F.~Sun, D.~Harwath, and J.~Glass.
\newblock Look, listen, and decode: Multimodal speech recognition with images.
\newblock In {\em IEEE Spoken Language Technology Workshop}, 2016.

\bibitem{Suris2018}
D.~Suris, A.~Duarte, A.~Salvador, J.~Torres, and X.~G. i~Nieto.
\newblock Cross-modal embeddings for video and audio retrieval.
\newblock In {\em arxiv:1801.02200}, 2018.

\bibitem{inceptionResnetv2}
C.~Szegedy, S.~Ioffe, V.~Vanhoucke, and A.~A. Alemi.
\newblock Inception-v4, inception-resnet and the impact of residual connections
  on learning.
\newblock In {\em Proceedings of the Thirty-First {AAAI} Conference on
  Artificial Intelligence, February 4-9, 2017, San Francisco, California,
  {USA.}}, pages 4278--4284, 2017.

\bibitem{PengPCL_TPAMI2018}
P.~Tang, X.~Wang, S.~Bai, W.~Shen, X.~Bai, W.~Liu, and A.~L. Yuille.
\newblock {PCL}: Proposal cluster learning for weakly supervised object
  detection.
\newblock {\em IEEE transactions on pattern analysis and machine intelligence},
  2018.

\bibitem{ss_ijcv}
J.~R. Uijlings, K.~E. Sande, T.~Gevers, and A.~W. Smeulders.
\newblock Selective search for object recognition.
\newblock {\em Int. J. Comput. Vision}, 104(2):154--171, Sept. 2013.

\bibitem{unsupervised_dec}
J.~Xie, R.~Girshick, and A.~Farhadi.
\newblock Unsupervised deep embedding for clustering analysis.
\newblock In {\em Proceedings of the 33rd International Conference on
  International Conference on Machine Learning - Volume 48}, ICML'16, pages
  478--487, 2016.

\bibitem{Yu2013}
H.~Yu and J.~M. Siskind.
\newblock Grounded language learning from video described with sentences.
\newblock In {\em ACL}, 2013.

\bibitem{ZhangECCV2016}
R.~Zhang, P.~Isola, and A.~A. Efros.
\newblock Colorful image colorization.
\newblock In {\em ECCV}, 2016.

\bibitem{Zhao2018}
H.~Zhao, C.~Gan, A.~Rouditchenko, C.~Vondrick, J.~McDermott, and A.~Torralba.
\newblock The sound of pixels.
\newblock In {\em ECCV}, 2018.

\end{thebibliography}
}

\end{document}